%% file: paper.tex
\pdfoutput=1

\documentclass[11pt]{article}

\usepackage[final]{acl}

\usepackage{times}
\usepackage{latexsym}

\usepackage[T1]{fontenc}

\usepackage[utf8]{inputenc}

\usepackage{microtype}

\usepackage{inconsolata}

\usepackage{graphicx}

\usepackage{booktabs}
\usepackage{multirow}

\usepackage[most]{tcolorbox}
\newtcolorbox{promptbox}[1][]{
    colback=gray!10,      
    colframe=gray!50,     
    boxrule=0.5mm,        
    arc=1mm,              
    boxsep=0mm,
    fontupper=\ttfamily\scriptsize,  
    width=0.45\textwidth,     
    title=#1,
    fonttitle=\ttfamily\footnotesize\centering,
}

\usepackage{stfloats}  

\title{Chain of Draft: Thinking Faster by Writing Less}

\makeatletter
\renewcommand{\@fnsymbol}[1]{$\dagger$}
\makeatother

\author{
  Silei Xu\thanks{Correspondence to <\texttt{silei.xu@zoom.us}>}, Wenhao Xie, Lingxiao Zhao, Pengcheng He\\
  Zoom Communications \\
}

\begin{document}
\maketitle
\begin{abstract}
Large Language Models (LLMs) have demonstrated remarkable performance in solving complex reasoning tasks through mechanisms like Chain-of-Thought (CoT) prompting, which emphasizes verbose, step-by-step reasoning. 
However, humans typically employ a more efficient strategy: drafting concise intermediate thoughts that capture only essential information. 
In this work, we propose {\em Chain of Draft} (CoD), a novel paradigm inspired by human cognitive processes, where LLMs generate minimalistic yet informative intermediate reasoning outputs while solving tasks. 
By reducing verbosity and focusing on critical insights, CoD matches or surpasses CoT in accuracy while using as little as only 7.6\% of the tokens, significantly reducing cost and latency across various reasoning tasks.
Our code and data are available at \url{https://github.com/sileix/chain-of-draft}.

\end{abstract}

\input{sections/intro}
\input{sections/related_work}
\input{sections/cod}

\input{sections/experiments}

\input{sections/conclusion}

\bibliography{custom}




\end{document}

%% file: sections/intro.tex
\section{Introduction}
Recent advances in reasoning models such as OpenAI o1~\cite{o1} and DeepSeek R1~\cite{r1} have propelled large language models (LLMs) to unprecedented performance on complex tasks using techniques like Chain of Thought (CoT)~\cite{cot}.
This paradigm encourages models to break down problems into step-by-step explorations, mimicking the structured reasoning process of humans. 
While effective, this approach demands substantially more computational resources at inference time, leading to verbose outputs and higher latency. 
Such verbosity contrasts sharply with how humans typically approach problem-solving: we rely on concise drafts or shorthand notes to capture essential insights without unnecessary elaboration.

Motivated by this difference, we propose Chain of Draft (CoD), a novel prompting strategy that aligns more closely with human reasoning by prioritizing efficiency and minimalism. Instead of verbose intermediate steps, Chain of Draft encourages LLMs to generate concise, dense-information outputs at each step. This approach reduces latency and computational costs without sacrifice of accuracy, making LLMs more practical for real-world applications where efficiency is paramount.

\begin{figure}[t]
    \vspace{-1.5em}
    \centering
    \includegraphics[width=0.4\textwidth]{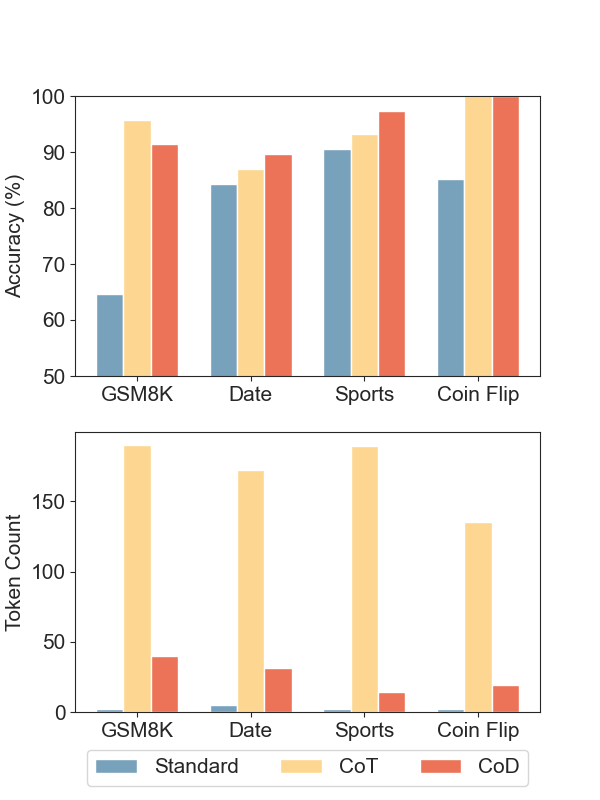}
    \caption{Comparison of Claude 3.5 Sonnet's accuracy and token usage across different tasks with three different prompt strategies: direct answer (Standard), Chain of Thought (CoT), and Chain of Draft (CoD). CoD achieves similar accuracy as CoT while using significant fewer tokens.}
    \label{fig:plot}
\end{figure}

The intuition behind Chain of Draft is rooted in how humans externalize thought. When solving complex tasks — whether solving mathematical problems, drafting essays, or coding — we often jot down only the critical pieces of information that help us progress. By emulating this behavior, LLMs can focus on advancing toward solutions without the overhead of verbose reasoning.

To evaluate the effectiveness of Chain of Draft, we conducted experiments across a variety of benchmarks requiring multi-step reasoning, including arithmetic reasoning, common sense reasoning, and symbolic reasoning. Our results demonstrate that this minimalist approach maintains or even improves accuracy compared with standard Chain of Thought, while significantly reducing token usage and latency. 

The contributions of this paper are threefold:
\begin{itemize}
    \item We introduce Chain of Draft, a concise reasoning prompting strategy inspired by human cognitive processes.
    \item We empirically validate that Chain of Draft can achieve significantly reduced latency and cost without sacrificing accuracy.
    \item We discuss the implications of Chain of Draft for LLM design, deployment, and real-world usability.
\end{itemize}



%% file: sections/related_work.tex
\section{Related Work}
\noindent{\bf Structured Reasoning Frameworks for LLMs}
Recently, a variety of reasoning language models have emerged, including o1 by OpenAI~\cite{o1}, QwQ by Alibaba~\cite{qwq}, and R1 by DeepSeek~\cite{r1}, demonstrating substantial improvements in tackling complex tasks. These models leverage structured reasoning methods to enhance robustness and problem-solving capabilities.
The concept of Chain-of-Thought reasoning (CoT)~\cite{cot, 0shot-cot}, established a foundational approach to reasoning in LLMs. 
Building on this foundation, more sophisticated topologies have emerged, such as tree~\cite{tot, bot, propagation} and graph~\cite{got, got2, resprompt}, enabling LLMs to address increasingly intricate problems.

Other enhancements include self-consistency CoT~\cite{self-consistency}, which incorporates verification and reflection mechanisms to bolster reasoning reliability, and ReAct~\cite{react}, which integrates tool usage into the reasoning process, allowing LLMs to access external resources and knowledge. These innovations collectively expand the reasoning capabilities of LLMs across a diverse range of applications.

\noindent{\bf LLM Inference Latency Reduction}
Although structured reasoning greatly enhances LLMs' ability to solve complex questions, 
it significantly increases the token usage before arriving at a final answer.
This makes it challenging to apply in cost-sensitive and latency-sensitive scenarios~\cite{token_economy}.
Furthermore, the model’s lack of awareness regarding task complexity often leads to overthinking~\cite{over-thinking, over-reasoning} even on simple tasks, resulting in unnecessary resource consumption.

Techniques like streaming aim to reduce {\em perceived} latency by incrementally providing partial outputs as they are generated, rather than waiting for the entire output sequence. However, this approach cannot fully mitigate overall latency or computational cost, and it is often unsuitable for chain-of-thought reasoning, as intermediate steps are often not intended to be shown to end users.

\citet{skeleton_of_thought} proposes Skeleton-of-Thought (SoT), a method that first guides LLMs to generate a skeleton outline of the answer, followed by parallel decoding to reduce latency. While SoT helps lower latency, it does not reduce computational cost and is limited to questions that can be parallelized effectively. 
\citet{draft_n_verify} took a different approach, it first generates draft tokens at lower quality but higher speed through selective skipping of intermediate layers, and then validates the draft in a single forward pass.
Our approach, CoD, can be combined with these approaches to further reduce the latency. 

\citet{latent-cot} proposes Coconut to train LLMs to perform reasoning in a continuous latent space rather than in the traditional natural language space using the final hidden state of the LLM to represent the reasoning process. 
While Coconut reduces latency and computational cost, it suffers from reduced accuracy in complex tasks, such as GSM8k. Additionally, it loses the interpretability of natural language reasoning and cannot be applied to black-box models like GPT and Claude.

The works closest to ours are Concise Thoughts (CCoT)~\cite{ccot} and token-budget-aware LLM reasoning (TALE)~\cite{budget}.
CCoT proposes using a fixed global token budget for reasoning steps. However, different tasks may require varying budgets to achieve the optimal balance between performance and cost. Moreover, LLMs may fail to adhere to an impractical budget, often generating far more tokens than intended~\cite{budget}.
\citet{budget} extends this idea by dynamically estimating a global token budget for different problems based on reasoning complexity. However, this approach requires an additional LLM call to estimate the budget, which increases latency. Furthermore, it assumes that the model can accurately predict the complexity of requests, limiting its applicability to more complex tasks where reflection, self-correction, or external knowledge retrieval may be necessary during the reasoning process. In contrast, our approach employs a per-step budget, allowing unlimited reasoning steps, which makes it more adaptable to various structured reasoning techniques.

%% file: sections/cod.tex
\section{Chain-of-Draft Prompting}

The Chain-of-Thought (CoT) prompting strategy has demonstrated significant effectiveness across a wide range of tasks, particularly those requiring complex multi-step reasoning.
However, LLMs often produce excessively verbose reasoning steps, consuming a substantial number of tokens before arriving at a final answer.
In contrast, humans tend to adopt a more concise approach when solving complex problems involving multi-step reasoning, such as mathematical or logical puzzles. Rather than elaborating on every detail, humans typically jot down only the essential intermediate results — minimal drafts — to facilitate their thought processes. Inspired by this natural tendency, we propose a novel prompting strategy called Chain-of-Draft (CoD). This approach aims to reduce verbosity by limiting the number of words used in each reasoning step, focusing only on the essential calculations or transformations needed to progress.

To illustrate the difference between standard prompting, Chain-of-Thought prompting, and our proposed Chain-of-Draft prompting, consider the following simple arithmetic problem:

\noindent\textit{Q: Jason had 20 lollipops. He gave Denny some lollipops. Now Jason has 12 lollipops. How many lollipops did Jason give to Denny?}

The response generated by a standard prompting approach directly outputs the answer, often without any reasoning. While correct, this lacks transparency in how the answer was derived, and requires the language model to run multi-step reasoning without any help from intermediate result, which often leads to hallucination.

\begin{center}
\begin{promptbox}[Standard]
Q: Jason had 20 lollipops. He gave Denny some lollipops. Now Jason has 12
lollipops. How many lollipops did Jason give to Denny?
\newline
A: 8
\end{promptbox}
\end{center}

Chain-of-Thought prompting, on the other hand, provides a detailed reasoning process.
While this response is accurate and interpretable, it includes unnecessary details about Jason, Denny, and the lollipops, which are irrelevant to solving the mathematical problem. This verbosity inflates the token count and increases the response latency.

\begin{center}
\begin{promptbox}[Chain-of-Thought]
Q: Jason had 20 lollipops. He gave Denny some lollipops. Now Jason has 12
lollipops. How many lollipops did Jason give to Denny?
\newline
A: Let's think through this step by step:

1. Initially, Jason had 20 lollipops. \newline
2. After giving some to Denny, Jason now has 12 lollipops. \newline
3. To find out how many lollipops Jason gave to Denny, we need to calculate the difference between the initial number of lollipops and the remaining number. \newline
4. We can set up a simple subtraction problem:
   Initial number of lollipops - Remaining number of lollipops = Lollipops given to Denny \newline
5. Putting in the numbers:
   20 - 12 = Lollipops given to Denny \newline
6. Solving the subtraction:
   20 - 12 = 8

Therefore, Jason gave 8 lollipops to Denny.

\#\#\#\#\newline
8 lollipops
\end{promptbox}
\end{center}

By contrast, Chain-of-Draft prompting condenses the reasoning process into minimal, abstract representations. Here, the reasoning is distilled into a concise equation, focusing solely on the essential mathematical operation required to arrive at the solution. By abstracting away irrelevant contextual details, CoD significantly reduces the token count while maintaining transparency and correctness.

\begin{center}
\begin{promptbox}[Chain-of-Draft]
Q: Jason had 20 lollipops. He gave Denny some lollipops. Now Jason has 12
lollipops. How many lollipops did Jason give to Denny?
\newline
A: 20 - x = 12; x = 20 - 12 = 8. \#\#\#\# 8
\end{promptbox}
\end{center}

%% file: sections/experiments.tex
\section{Experiments}
In empirical experiments, we follow the original CoT paper~\cite{cot} to evaluate on 3 categories of tasks: arithmetic reasoning, commonsense reasoning, and symbolic reasoning. 
We pick representative tasks where original CoT significantly improves the accuracy over the baseline without reasoning. In particular, we pick GSM8k~\cite{gsm8k} for arithmetic reasoning; date understanding and sports understanding from BIG-bench~\cite{bigbench} for commonsense reasoning; and coin flip tasks introduced in the CoT paper~\cite{cot} for symbolic reasoning. 

\subsection{Experimental Setup}
We compare three different prompt strategies: CoT, CoD, and Standard prompting as a baseline.

\noindent{\bf Standard prompting.} we use standard few-shot prompting~\cite{fewshot}, where the model is given input-output pairs as in-context examples. LLMs are asked to directly return the final answer, without any reasoning or explanation. 

\noindent{\bf Chain-of-Thought.}
We follow the exact few-shot examples provided in the appendix of the CoT paper with the exception of having the final answer after four hashtags ({\small \#\#\#\#}) for a more stable answer extraction. 

\noindent{\bf Chain-of-Draft.} In CoD, we also asked the model to think step by step. However, the model is asked to limit each reasoning step to five words at most. Note that we do not enforce such limitation in any way, it is just a general guideline to promte short reasoning steps. 
For each few-shot example, we also include the Chain of Draft written manually by the authors. 

The complete system prompt for each prompting strategy is shown below.

\begin{center}
\centering
\begin{promptbox}[Standard]
Answer the question directly.
Do not return any preamble, explanation, or reasoning.
\end{promptbox}
\begin{promptbox}[Chain-of-Thought]
Think step by step to answer the following question.
Return the answer at the end of the response after a separator \#\#\#\#.
\end{promptbox}
\begin{promptbox}[Chain-of-Draft]
Think step by step, but only keep a minimum draft for each thinking step, with 5 words at most.
Return the answer at the end of the response after a separator \#\#\#\#.
\end{promptbox}
\end{center}

We evaluated each task with two of the most popular flagship models: GPT-4o (gpt-4o-2024-08-06) from OpenAI and Claude 3.5 Sonnet (claude-3-5-sonnet-20240620) from Anthropic.

\subsection{Arithmetic Reasoning}
We first consider math problems that measure the arithmetic reasoning capabilities of LLMs.  
GSM8k~\cite{gsm8k} has emerged as the benchmark of choice for evaluating arithmetic reasoning in language models, providing a comprehensive dataset of 8,500 diverse grade-school-level mathematical problems. Each problem is paired with a detailed step-by-step solution, emphasizing arithmetic, geometry, algebra, and logical reasoning skills. 

The evaluation results are presented in Table~\ref{tab:gsm8k}. The dataset poses significant challenges for both GPT-4o and Claude 3.5 Sonnet when using standard prompting, yielding accuracies of 53.3\% and 64.6\%, respectively. However, with the application of the CoT, both models surpass 95\% accuracy, albeit at the expense of generating approximately 200 tokens per response. In contrast, CoD achieves an accuracy of 91\% for both models while requiring only about 40 tokens per response, thereby reducing the average output token count by 80\% and cutting the average latency by 76.2\% and 48.4\%, respectively.

\begin{table}[!ht]
\centering
\fontsize{8.5}{8}\selectfont
\begin{tabular}{llrrr}
\toprule
\textbf{Model} & \textbf{Prompt} & \textbf{Accuracy} & \textbf{Token \#} & \textbf{Latency}\\
\midrule
\multirow{4}{*}{GPT-4o} & Standard & 53.3\% & 1.1 & 0.6 s\\
\cmidrule{2-5}
& CoT & 95.4\% & 205.1 & 4.2 s\\
\cmidrule{2-5}
& CoD & 91.1\% & 43.9 & 1.0 s\\
\midrule
\multirow{4}{*}{\shortstack[l]{Claude 3.5 \\Sonnet}} & Standard & 64.6\% & 1.1 & 0.9 s \\
\cmidrule{2-5}
& CoT & 95.8\% & 190.0 & 3.1 s\\
\cmidrule{2-5}
& CoD & 91.4\% & 39.8 & 1.6 s\\
\bottomrule
\end{tabular}
\caption{GSM8K evaluation results.}
\label{tab:gsm8k}
\end{table}

\subsection{Commonsense Reasoning}
We evaluate the tasks of date understanding and sports understanding from BIG-bench to demonstrate the effectiveness of CoD in common sense reasoning. For consistency, we use the same system prompts as those employed in the arithmetic reasoning evaluation.

The evaluation results, presented in Table~\ref{tab:date_understanding}, show that CoD significantly reduces both latency and cost by generating considerably fewer tokens in responses compared to CoT. Additionally, CoD outperforms CoT in accuracy in various cases. 
Notably, chain-of-thought prompting leads to excessively verbose responses for Claude 3.5 Sonnet, especially in the sports understanding task, where CoD reduces the average output tokens from 189.4 to 14.3 — a 92.4\% reduction.

\begin{table}[!ht]
\centering
\fontsize{8}{8}\selectfont
\begin{tabular}{llrrr}
\toprule
\textbf{Model} & \textbf{Prompt} & \textbf{Accuracy} & \textbf{Token \#} & \textbf{Latency}\\
\midrule
\multirow{4}{*}{GPT-4o} & Standard & 72.6\% & 5.2 & 0.6 s\\
\cmidrule{2-5}
& CoT & 90.2\% & 75.7 & 1.7 s\\
\cmidrule{2-5}
& CoD & 88.1\% & 30.2 & 1.3 s\\
\midrule
\multirow{4}{*}{\shortstack[l]{Claude 3.5 \\Sonnet}} & Standard & 84.3\% & 5.2 & 1.0 s \\
\cmidrule{2-5}
& CoT & 87.0\% & 172.5 & 3.2 s\\
\cmidrule{2-5}
& CoD & 89.7\% & 31.3 & 1.4 s\\
\bottomrule
\end{tabular}
\caption{Date understanding evaluation results.}
\label{tab:date_understanding}
\end{table}

\begin{table}[!ht]
\centering
\fontsize{8}{8}\selectfont
\begin{tabular}{llrrr}
\toprule
\textbf{Model} & \textbf{Prompt} & \textbf{Accuracy} & \textbf{Token \#} & \textbf{Latency}\\
\midrule
\multirow{4}{*}{GPT-4o} & Standard & 90.0\% & 1.0 & 0.4 s\\
\cmidrule{2-5}
& CoT & 95.9\% & 28.7 & 0.9 s\\
\cmidrule{2-5}
& CoD & 98.3\% & 15.0 & 0.7 s\\
\midrule
\multirow{4}{*}{\shortstack[l]{Claude 3.5 \\Sonnet}} & Standard & 90.6\% & 1.0 & 0.9 s \\
\cmidrule{2-5}
& CoT & 93.2\% & 189.4 & 3.6 s\\
\cmidrule{2-5}
& CoD & 97.3\% & 14.3 & 1.0 s\\
\bottomrule
\end{tabular}
\caption{Sports understanding evaluation results.}
\label{tab:sports_understanding}
\end{table}

\subsection{Symbolic Reasoning}
The original CoT paper~\cite{cot} introduces the task of coin flipping,
where the LLMs are asked to predict which side is up after a sequence of coin flip actions.
Since the exact dataset is not published, we synthesize a test set of 250 examples following the same design.
Specifically, we randomly chose 4 out of the top 1000 first names in the US region according to NameDataset~\cite{name_dataset} and randomly decided to flip the coin or not for each name. 
An example of the evaluation data is shown below. 

\begin{center}
\begin{promptbox}
Q: A coin is heads up. Robyn flips the coin. Peggy flips the coin. Grant flips the coin. Vanessa does not flip the coin. Is the coin still heads up?
\newline
A: No.
\end{promptbox}
\end{center}

The evaluation results for GPT-4o and Claude 3.5 Sonnet are shown in Table~\ref{tab:coin_flip}. They achieve 73.2\% and 85.2\% with standard prompting, respectively. However, both models reach a perfect 100\% accuracy with CoT and CoD. Again, CoD demonstrates significant reduction of tokens compared to CoT, from 68\% for GPT-4o to 86\% for Claude 3.5 Sonnet.

\begin{table}[!ht]
\centering
\fontsize{8}{8}\selectfont
\begin{tabular}{llrrr}
\toprule
\textbf{Model} & \textbf{Prompt} & \textbf{Accuracy} & \textbf{Token \#} & \textbf{Latency}\\
\midrule
\multirow{4}{*}{GPT-4o} & Standard & 73.2\% & 1.0 & 0.4 s\\
\cmidrule{2-5}
& CoT & 100.0\% & 52.4 & 1.4 s\\
\cmidrule{2-5}
& CoD & 100.0\% & 16.8 & 0.8 s\\
\midrule
\multirow{4}{*}{\shortstack[l]{Claude 3.5 \\Sonnet}} & Standard & 85.2\% & 1.0 & 1.2 s \\
\cmidrule{2-5}
& CoT & 100.0\% & 135.3 & 3.1 s\\
\cmidrule{2-5}
& CoD & 100.0\% & 18.9 & 1.6 s\\
\bottomrule
\end{tabular}
\caption{Coin flip evaluation results.}
\label{tab:coin_flip}
\end{table}

\subsection{Limitaitons of CoD}

\noindent {\bf Inconsistency Without Few-shot Examples}

\noindent We evaluated the performance of CoD under zero-shot setting, where no few-shot examples were provided. The results, presented in Table~\ref{tab:zero_shot}, indicate a significant decline in CoD's effectiveness. Notably, for Claude 3.5 Sonnet, CoD improved performance over direct answering by only 3.6\%. Additionally, the token savings achieved by CoD are less significant compared to few-shot setting.

We hypothesize that this limitation arises due to the scarcity or absence of CoD-style reasoning patterns in the training data of large language models, making it a challenging task to generate concise and insightful ``drafts'' without guidance from few-shot examples. 

\begin{table}[!ht]
\centering
\fontsize{8.5}{8}\selectfont
\begin{tabular}{llrrr}
\toprule
\textbf{Model} & \textbf{Prompt} & \textbf{Accuracy} & \textbf{Token \#} & \textbf{Latency}\\
\midrule
\multirow{4}{*}{GPT-4o} & Standard & 56.9\% & 2.2 & 0.5 s\\
\cmidrule{2-5}
& CoT & 94.8\% & 278.4 & 8.1 s\\
\cmidrule{2-5}
& CoD & 84.4\% & 76.4 & 2.6 s\\
\midrule
\multirow{4}{*}{\shortstack[l]{Claude 3.5 \\Sonnet}} & Standard & 61.9\% & 5.2 & 0.9 s \\
\cmidrule{2-5}
& CoT & 90.4\% & 248.8 & 3.5 s\\
\cmidrule{2-5}
& CoD & 65.5\% & 73.7 & 1.6 s\\
\bottomrule
\end{tabular}
\caption{Zero-shot GSM8K evaluation results.}
\label{tab:zero_shot}
\end{table}

\noindent {\bf Reduced Performance on Small Models}

\noindent We tested CoD on several small language models with fewer than 3B parameters, including Qwen2.5 1.5B/3B instruct~\cite{qwen25}, Llama 3.2 3B instruct~\cite{llama3}, and our in-house Zoom SLM 2.3B model~\cite{zoom-slm}. While CoD effectively reduces the number of tokens required per response and improves accuracy over direct answer, its performance gap compared to CoT is more pronounced in these models. 

Similar to the zero-shot setting, we suspect this is due to the absence of CoD-style data in the training process. We anticipate that fine-tuning these models with additional CoD-formatted data could significantly enhance their reasoning accuracy with CoD.

\begin{table}[!ht]
\centering
\fontsize{8.5}{8}\selectfont
\begin{tabular}{llrr}
\toprule
\textbf{Model} & \textbf{Prompt} & \textbf{Accuracy} & \textbf{Token \#}\\
\midrule
\multirow{4}{*}{Qwen2.5-1.5B-Instruct} & Standard & 5.7\% & 6.6 \\
\cmidrule{2-4}
& CoT & 32.5\% & 141.4 \\
\cmidrule{2-4}
& CoD & 24.2\% & 75.1\\
\midrule
\multirow{4}{*}{Qwen2.5-3B-Instruct} & Standard & 7.2\% & 3.4 \\
\cmidrule{2-4}
& CoT & 59.1\% & 236.4 \\
\cmidrule{2-4}
& CoD & 43.1\% & 41.2\\
\midrule
\multirow{4}{*}{Llama3.2-3B-Instruct} & Standard & 3.9\% & 16.6  \\
\cmidrule{2-4}
& CoT & 70.7\% & 195.3 \\
\cmidrule{2-4}
& CoD & 52.5\% & 98.1\\
\midrule
\multirow{4}{*}{\shortstack[l]{Zoom-SLM-2.3B}} & Standard & 5.9\% & 3.8  \\
\cmidrule{2-4}
& CoT & 77.7\% & 129.0 \\
\cmidrule{2-4}
& CoD & 50.9\% & 55.6 \\
\bottomrule
\end{tabular}
\caption{GSM8K evaluation results on small language models.}
\label{tab:slm}
\end{table}


%% file: sections/conclusion.tex
\section{Discussion}
The latency issue has often been overlooked in studies of the reasoning capabilities of LLMs. 
However, it is crucial for lots of real-time applications to have low latency while maintaining high-quality responses. 
In this work, we propose Chain of Draft (CoD), a novel approach that substantially reduces the latency required for reasoning while achieving comparable or even superior accuracy compared to standard Chain-of-Thought prompting strategies. Unlike traditional methods that often involve lengthy reasoning steps, CoD leverages concise reasoning drafts to speed up response generation without sacrificing correctness.

Additionally, CoD offers significant cost advantages. By compacting the reasoning steps, it reduces the number of input tokens required for few-shot prompting and shortens the output token length, directly lowering computational cost. This token efficiency makes CoD especially appealing in cost-sensitive scenarios, such as large-scale deployments of LLMs or applications with strict budget constraints.

CoD demonstrates that {\bf\em effective reasoning in LLMs does not necessarily require lengthy outputs}, offering an alternative approach where reasoning depth is maintained with minimal verbosity. 
Future work could explore combining CoD with other latency-reducing methods, such as adaptive parallel reasoning or multi-pass validation, to further optimize performance across different application domains. In addition, the principles behind the compact reasoning of CoD could inspire new strategies to improve reasoning models by training with compact reasoning data, while maintaining interpretability and efficiency in LLMs, helping bridge the gap between research-driven improvements in reasoning and the practical demands of real world systems.